\documentclass[conference]{IEEEtran}
\IEEEoverridecommandlockouts
\usepackage{cite}
\usepackage{amsmath,amssymb,amsfonts}
\usepackage{algorithmic}
\usepackage{graphicx}
\graphicspath{{./}}
\usepackage{textcomp}
\usepackage{xcolor}
\usepackage{booktabs}
\usepackage{multirow}
\usepackage{url}

\def\BibTeX{{\rm B\kern-.05em{\sc i\kern-.025em b}\kern-.08em
    T\kern-.1667em\lower.7ex\hbox{E}\kern-.125emX}}

\begin{document}

\title{ARID: A Deployable Edge AI System for Structured Information Extraction from Industrial Maintenance Work Orders\\
\thanks{\copyright~2026 IEEE. Personal use of this material is permitted. Permission from IEEE must be obtained for all other uses, in any current or future media, including reprinting/republishing this material for advertising or promotional purposes, creating new collective works, for resale or redistribution to servers or lists, or reuse of any copyrighted component of this work in other works. Accepted for publication at IEEE IECON 2026.}}

\author{
\IEEEauthorblockN{Kuanlin Chen}
\IEEEauthorblockA{\textit{Independent Researcher}\\
Taoyuan, Taiwan \\
dipper13579@gmail.com \\
ORCID: 0009-0006-3879-4565}
\and
\IEEEauthorblockN{Chen-Wei Kuo}
\IEEEauthorblockA{\textit{National Tsing Hua University}\\
Hsinchu, Taiwan \\
ck3294@nyu.edu}}

\maketitle

\begin{abstract}
Maintenance work orders must often be processed offline on embedded hardware, yet downstream software requires predictable structured output. We present ARID (Aviation-inspired Routing for Industrial Deployment), which extracts component, failure mode, symptom, and maintenance action into fixed-schema JSON on an 8\,GB NVIDIA Jetson Orin NX. ARID combines conservative dual-teacher filtering, targeted noise-aware synthesis, one routing decision per work order, 4-bit inference, and grammar-constrained decoding. From 2,326 unlabeled OMIn records, it retains 716 training pairs and adds 99 topology-constrained records targeting action extraction. On 300 human-labeled records, ARID reaches 84.8\% token-F1 on the reference stack and 82.9\% on the deployed Jetson. Resident serving achieves 5{,}310/5{,}656\,ms P50/P99 at 12.5\,W. On zero-shot MaintNet transfer, semantic F1 falls to 46.4\% while parser success remains at least 99.8\%, showing that output validity transfers but field semantics do not.
\end{abstract}

\begin{IEEEkeywords}
Industrial Informatics, Edge AI, Maintenance Work Orders, Large Language Models, System Deployment, NVIDIA Jetson
\end{IEEEkeywords}

\section{Introduction}

Technician narratives contain the evidence needed to populate maintenance databases, but they are short, noisy, and domain-specific. The operational task is therefore stricter than summarization: given one work order, a system must recover four fields---\textit{component}, \textit{failure mode}, \textit{symptom}, and \textit{maintenance action}---and return a record that a CMMS can validate deterministically.

Factory deployment adds three coupled constraints. Logs may be proprietary, the target network may be air-gapped, and an embedded device must provide predictable resident-serving behavior within an 8\,GB shared-memory budget. At the same time, high-quality field annotations are scarce. Prior maintenance pipelines address extraction or reasoning \cite{keo2025}, and prior edge work addresses model execution, but neither result alone establishes a complete path from noisy text to validated records on the device.

ARID addresses this gap as a hardware-aware pipeline rather than as a new foundation model. Offline, two independent teachers propose structured labels; a conservative agreement filter and targeted Noise-Aware Synthetic Distillation (NASD) construct the fine-tuning set. Online, a semantic gate chooses once per work order whether to apply the domain adapter, a 4-bit model performs extraction, and Grammar-Constrained Decoding (GCD) enforces the JSON grammar. This separation is deliberate: training data affects semantic accuracy, routing determines which specialization is used, and GCD guarantees syntax but cannot repair a wrong field value.

The evaluation makes the same separation. ARID obtains 84.8\% token-F1 on the RTX reference stack and retains 82.9\% after deployment conversion to llama.cpp Q4\_K\_M on the Jetson. Resident serving reaches 5{,}310/5{,}656\,ms P50/P99 at 12.5\,W. On MaintNet, parser success remains at least 99.8\% while semantic F1 falls to 46.4\%, exposing target-domain vocabulary---not JSON formation---as the transfer bottleneck.

The contributions are:
\begin{itemize}\setlength{\itemsep}{1pt}\setlength{\parsep}{0pt}\setlength{\topsep}{3pt}
    \item \textbf{A complete measured edge pipeline:} ARID connects low-label training, work-order-level adapter selection, 4-bit inference, and constrained JSON generation on a commercial Jetson Orin NX~8GB.
    \item \textbf{Targeted low-resource data construction:} dual-teacher filtering yields 716 retained pairs, while 99 topology-constrained NASD records produce a significant 1.56\,pp action-field gain in the $N=100$ setting; we distinguish this targeted effect from the nonsignificant overall gain.
    \item \textbf{A semantic--structural transfer diagnosis:} cross-domain evaluation shows that constrained decoding preserves machine validity under shift even when field extraction fails, identifying vocabulary adaptation as the next deployment requirement.
\end{itemize}

\begin{figure*}[t]
  \centering
  \includegraphics[width=\textwidth]{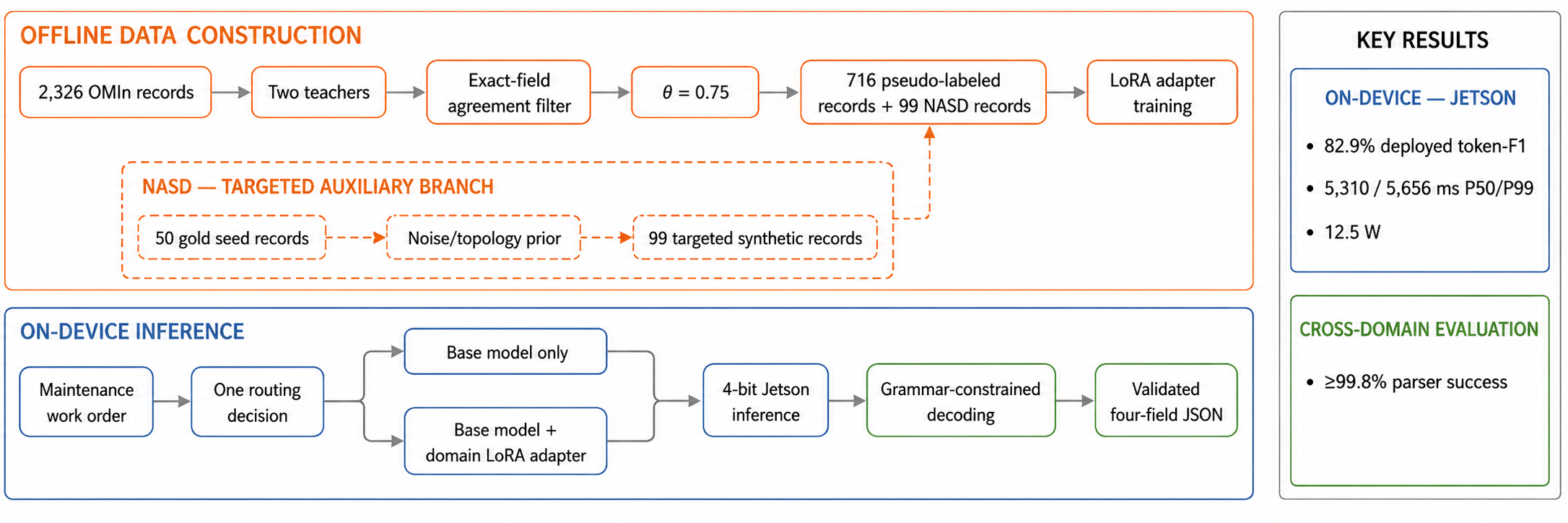}
  \caption{ARID separates the source of semantic supervision (top) from the deployed extraction path (bottom). The router selects specialization once per record; GCD constrains the JSON grammar but does not determine field content.}
  \label{fig:overview}
\end{figure*}

\section{Related Work}

\textbf{Industrial maintenance IE.}
Extraction from maintenance narratives has progressed from statistical classifiers~\cite{rahman2023classification} to resources spanning aviation, automotive, and facility domains. MaintNet~\cite{akhbardeh2020maintnet} provides a collaborative maintenance library; MaintIE~\cite{french2024maintie} defines a fine-grained 224-class schema. OMIn~\cite{omin2025trusted} contains FAA-derived incident records and supports knowledge-extraction pipelines such as KEO~\cite{keo2025}. These works establish the semantic task; ARID asks what is required to execute a smaller four-field schema locally and return validated records under an embedded-device budget.

\textbf{Edge LLM deployment.}
Four-bit quantization reduces the dense-weight footprint of small LLMs~\cite{sparrenberg2025small,kurt2026unified} and enables execution on constrained devices~\cite{arya2025edge}. Jetson studies characterize model-scale, latency, and power trade-offs~\cite{arya2025edge}; EdgeLoRA~\cite{edgelora2025} and CLONE~\cite{clone2025} address multi-adapter serving and latency-aware scheduling. ARID builds on these deployment mechanisms but evaluates an end-to-end maintenance extractor, including output validity and cross-domain behavior.

\textbf{Mixture-of-LoRA adapter routing.}
LoRA~\cite{hu2021lora} has been extended to token-level mixtures such as MoLE~\cite{mole2024}, MoLoRA~\cite{molora2025}, and LD-MoLE~\cite{ldmole2025}; EdgeMoE~\cite{edgemoe2023} separates hot backbone weights from cold expert weights. ARID does not route tokens or compose experts. It makes one auditable base-versus-adapter decision before generation, trading routing expressiveness for a simpler edge execution path.

\textbf{Structured decoding and annotation bootstrapping.}
GCD masks illegal tokens according to a finite-state grammar~\cite{willard2023automata}; it has improved structured IE in low-resource settings~\cite{schmidt2025gcd}. Separately, knowledge-graph-grounded prompting can generate realistic maintenance orders~\cite{generating_authentic2025}, and Noise-Aware Training improves robustness to corrupted text~\cite{nat2020}. ARID combines these ideas but assigns them different roles: synthesis targets semantic coverage, whereas GCD enforces structure only.

\section{ARID System Design}

Fig.~\ref{fig:overview} separates offline data construction from the on-device path. A conventional 3B model requires roughly 6\,GB for FP16 weights before KV cache and runtime buffers; ARID therefore uses a 4-bit backbone and a compact LoRA adapter. The output contract is a four-key JSON object. The design goal is not to make every field correct by construction---a grammar cannot do that---but to make semantic errors measurable and syntactic failures preventable.

\subsection{Agreement-Filtered Pseudo-Labels}
\begin{figure}[t]
  \centering
  \includegraphics[width=.92\columnwidth]{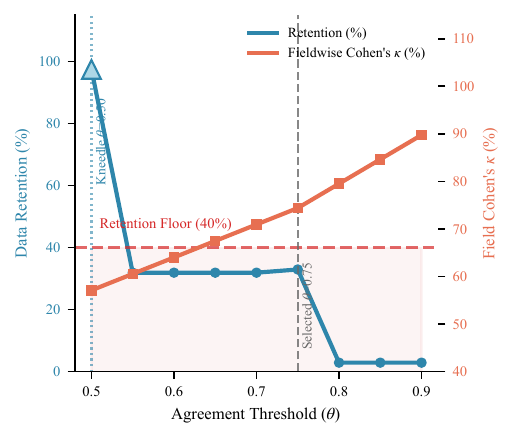}
  \caption{Teacher-agreement filtering. Kneedle locates the diagnostic knee at $\theta{=}0.50$; the stricter record-agreement threshold $\theta{=}0.75$ retains 716 records.}
  \label{fig:kappa}
\end{figure}

Two independent teachers, Gemini-2.5-Pro and DeepSeek-Chat, map each of 2,326 unlabeled OMIn work orders to the target schema. For record $i$, exact field agreement is
\begin{equation}
a_i=\frac{1}{4}\sum_{f=1}^{4}\mathbb{1}\!\left[y^{(1)}_{if}=y^{(2)}_{if}\right].
\end{equation}
For each threshold, records with $a_i\geq\theta$ are retained and fieldwise Cohen's $\kappa$ is computed over that retained set. Fig.~\ref{fig:kappa} shows the resulting agreement--retention curve. Kneedle~\cite{kneedle2011} identifies $\theta{=}0.50$ as the curve knee under a 40\% retention floor. We use this knee as a diagnostic, then choose the stricter $\theta{=}0.75$ because the deployment objective prioritizes label precision over corpus size. This requires exact teacher agreement on at least three of four fields and retains 716 records. The retained subset has fieldwise Cohen's $\kappa=0.744$. It is a conservative engineering override, not an automatically chosen optimum. The 40\% floor constrains the automatic knee search only; the selected $\theta{=}0.75$ retains 30.8\% and therefore lies below it by design. The seed, pseudo-label, and test pools are disjoint.

\subsection{Noise-Aware Synthetic Distillation (NASD)}
The pseudo-labeled set underrepresents the surface diversity of \textit{maintenance action}. NASD is therefore a targeted augmentation, not a replacement for pseudo-labeling. From a disjoint seed of 50 gold records, it derives a topology prior over components, actions, abbreviations, phoneme-confusion typos, and telegraphic grammar. A generator then produces 99 topology-constrained work orders with explicit action nodes. These records are added to the 716 pseudo-labels for adapter training. Section~\ref{sec:lowresource} evaluates the intended action-field effect separately from overall F1.

\subsection{Work-Order-Level Adapter Selection}
The input is embedded with \texttt{all-MiniLM-L6-v2} and compared with precomputed general and expert prototypes. If expert similarity exceeds the fixed development-set threshold $\tau=0.25$, ARID applies the domain LoRA adapter; otherwise it uses the shared base model. The decision is made once before generation, so it adds no per-token routing or adapter-composition loop.

At the selected operating point, 97.3\% of OMIn test records use the expert path. Consequently, the current router should be interpreted as an auditable specialization policy, not as evidence of large average compute savings on this domain. In resident serving, the adapter remains in memory; the cold swap experiment in Section~\ref{sec:deployment} quantifies the alternative disk-load path.

\subsection{Grammar-Constrained Decoding (GCD)}
The four-field schema is compiled into a finite-state machine that masks tokens violating the JSON grammar. GCD therefore guarantees character-level grammar validity, while a deterministic parser verifies that all required keys are present. The distinction matters: GCD prevents malformed payloads but cannot correct a semantically wrong component or action. In the on-device check ($N=10$), its median overhead is 1\,ms; across the full evaluations, grammar validity is 100\% and unseen-domain parser success is at least 99.8\%.

\begin{figure}[t]
  \centering
  \includegraphics[width=.92\columnwidth]{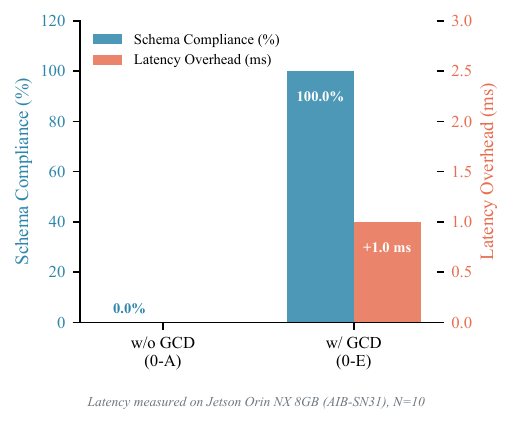}
  \caption{GCD on the Jetson Orin NX~8GB: 100\% character-level JSON grammar validity at 1\,ms median overhead; unconstrained zero-shot schema compliance is 0\% ($N=10$).}
  \label{fig:gcd}
\end{figure}

\section{Experiments and Hardware Validation}

\subsection{Experimental Setup}
\textbf{Evaluation backends.}
The reference stack uses bitsandbytes 4-bit quantization and outlines~1.2.12 on an RTX~3060~Ti (12\,GB). The deployed stack uses llama.cpp Q4\_K\_M on an Aetina AIB-SN31 with a Jetson Orin NX~8GB, JetPack~6.2, and the 20\,W power mode. We report the backend beside every accuracy value and reserve P50/P99 latency and power claims for measurements on the Jetson.

\textbf{Model and training configuration.}
All configurations use Qwen2.5-3B as the backbone. Adapters are trained with QLoRA (4-bit NF4 quantization, double quantization, bfloat16 compute), LoRA rank~8, $\alpha=16$, dropout~0.05, applied to \texttt{q\_proj} and \texttt{v\_proj}. Training uses 3 epochs, effective batch size~16 (per-device~4 with gradient accumulation~4), learning rate $2\times10^{-4}$, and the AdamW optimizer. The same recipe is used for every adapter reported in this paper.

The in-domain test set contains 300 human-annotated OMIn records. Token-F1 is micro-averaged over tokens in the four output fields. \textit{Grammar validity} means that the character sequence obeys the compiled grammar; \textit{parser success} additionally requires a JSON object with all four keys. The routing threshold is fixed at $\tau=0.25$ from the development sweep and is not retuned on either test domain.

\subsection{Main Results}

Table~\ref{tab:main} presents the accuracy chain without mixing reference and deployment backends in one cell. Pseudo-labeling supplies the largest training-data gain: 84.7\% versus 77.5\% for plain synthetic SFT. Adding NASD changes full-data F1 by only 0.1\,pp; its narrower low-resource action-field effect is tested next. Conversion from the RTX reference stack to the deployed Jetson stack costs 1.9\,pp, yielding 82.9\% token-F1 on device.

\begin{table}[t]
\caption{OMIn extraction accuracy ($N=300$). Backend is shown explicitly.}
\label{tab:main}
\centering
\setlength{\tabcolsep}{3.5pt}
\renewcommand{\arraystretch}{0.95}
\begin{tabular}{lccc}
\toprule
\textbf{Configuration} & \textbf{Backend} & \textbf{Token-F1} & \textbf{Grammar} \\
\midrule
Zero-shot, no GCD       & RTX    & 0.0\%$^\dagger$ & 0\% \\
Plain synthetic SFT     & RTX    & 77.5\% & 100\% \\
Pseudo-label SFT        & RTX    & 84.7\% & 100\% \\
\textbf{+ NASD (ARID reference)} & \textbf{RTX} & \textbf{84.8\%} & 100\% \\
\textbf{ARID deployed} & \textbf{Jetson} & \textbf{82.9\%} & 100\% \\
Cloud API zero-shot & Cloud & 67.3\% & 100\%$^\P$ \\
\bottomrule
\end{tabular}
\smallskip\\
\raggedright\footnotesize
RTX: bitsandbytes 4-bit + outlines. Jetson: llama.cpp Q4\_K\_M. The deployed row uses the fixed router ($\tau=0.25$, expert activation 97.3\%).\\
$^\dagger$Without fine-tuning or GCD, the backbone produced no parseable four-field JSON; all fields were therefore scored as missing.\\
$^\P$OpenAI GPT-5 Nano (\texttt{gpt-5-nano-2025-08-07}), evaluated via API on March~21, 2026. Schema compliance comes from prompt-based instruction following; no GCD was applied. This row is a deployment-contrast reference, not a same-supervision upper bound. External API latency is omitted because connectivity-dependent estimates are not commensurate with on-device measurements.
\end{table}

\vspace{-2pt}
\subsection{Low-Resource Ablation}\label{sec:lowresource}

To isolate NASD's intended effect, both low-resource models use the same $N=100$ pseudo-labeled subset; one additionally receives the 99 NASD records. The overall +0.4\,pp change is not significant (Wilcoxon $p=0.086$), whereas action extraction improves by 1.56\,pp (McNemar $p<0.001$). Thus NASD is supported as a targeted action-field intervention under scarcity, not as the source of ARID's overall 84.8\% result.

\begin{table}[t]
\caption{Low-Resource Ablation: NASD vs.\ Pseudo-Label Only ($N=100$)}
\label{tab:lowdata}
\centering
\setlength{\tabcolsep}{3.5pt}
\renewcommand{\arraystretch}{0.95}
\begin{tabular}{lcc}
\toprule
\textbf{Training Data} & \textbf{Token-F1} & \textbf{Action F1} \\
\midrule
100 pseudo only        & 81.2\% & 26.9\% \\
99 NASD + 100 pseudo   & \textbf{81.6\%} & \textbf{28.5\%} \\
\midrule
Difference            & +0.4\,pp$^\S$ & +1.56\,pp$^{**}$ \\
\bottomrule
\end{tabular}
\smallskip\\
\raggedright\footnotesize
$^\S$Not significant (Wilcoxon $p=0.086$).\\
$^{**}$Significant (McNemar $p<0.001$).\\
Wilcoxon tests are conducted over per-sample token-F1 distributions; McNemar tests are applied to paired binary correctness at the field level.
\end{table}

\vspace{-2pt}
\subsection{Latency and Hardware Validation}\label{sec:deployment}

All latency measurements use batch size 1, greedy decoding (temperature~=~0), and a fixed output budget of 80 tokens. For resident serving, model weights, adapter, and grammar compiler remain in memory via llama-server. $N=300$ test samples are evaluated sequentially.

Table~\ref{tab:latency} reports only measured Jetson configurations. External API latency is omitted because connectivity-dependent estimates are not commensurate with on-device measurements.

\begin{table}[t]
\caption{Measured Jetson latency ($N=300$, batch size 1).}
\label{tab:latency}
\centering
\setlength{\tabcolsep}{3pt}
\renewcommand{\arraystretch}{0.95}
\begin{tabular}{@{}lccc@{}}
\toprule
\textbf{Execution path} & \textbf{P50} & \textbf{P99} & \textbf{Disk-load cost} \\
\midrule
Resident serving & 5,310\,ms & 5,656\,ms & 0\,ms \\
Cold swap diagnostic & 6,805\,ms & 11,481\,ms & 1,495\,ms \\
\bottomrule
\end{tabular}
\smallskip\\
\raggedright\footnotesize
Jetson Orin NX~8GB, llama.cpp Q4\_K\_M, 20\,W mode. Resident P50/P99 exclude the first model load; observed maximum after that exclusion is 11,865\,ms.
\end{table}

\begin{figure}[t]
  \centering
  \includegraphics[width=.95\columnwidth]{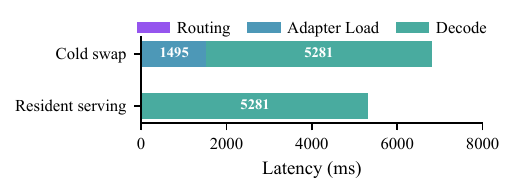}
  \caption{Inference latency breakdown on the Jetson Orin NX~8GB (AIB-SN31, 20\,W mode). Resident serving retains the adapter in memory via llama-server; cold swap reloads it from disk via llama-cli, adding 1,495\,ms.}
  \label{fig:latency}
\end{figure}

Fig.~\ref{fig:latency} shows why residency matters: loading the adapter from disk adds 1,495\,ms to the median path and broadens the tail. ARID therefore uses llama-server with the backbone, adapter, and grammar resident for normal operation; cold swap is retained only as a diagnostic configuration.

\begin{figure}[t]
  \centering
  \includegraphics[width=.95\columnwidth]{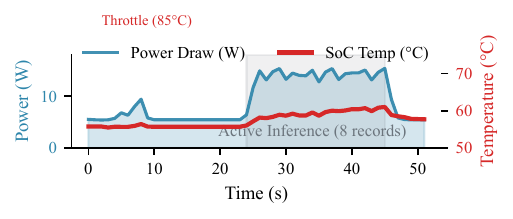}
  \caption{SoC power draw and temperature on the Aetina AIB-SN31 (20\,W mode). Active inference draws 12.5\,W; the system remains within thermal limits without throttling.}
  \label{fig:power}
\end{figure}

During sustained inference, the device draws 12.5\,W without thermal throttling (Fig.~\ref{fig:power}). Together with the measured latency distribution, this establishes a reproducible resident-serving operating point; it is not a hard real-time worst-case guarantee.

\vspace{-2pt}
\subsection{Domain Generalization}

\begin{figure}[t]
  \centering
  \includegraphics[width=.94\columnwidth]{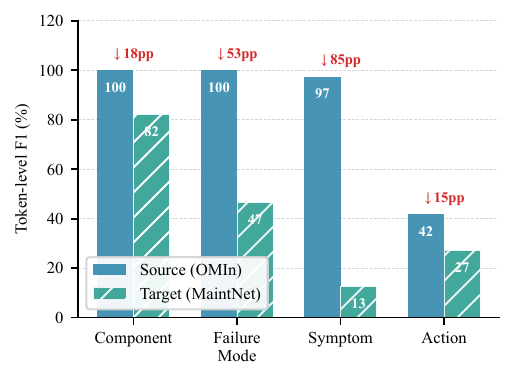}
  \caption{Per-field token-F1 on OMIn and MaintNet. \textit{Component} is more domain-generic; \textit{Symptom} exposes the vocabulary bottleneck. Parser success remains $\geq$99.8\% under GCD.}
  \label{fig:transfer}
\end{figure}

ARID is applied zero-shot to MaintNet~\cite{akhbardeh2020maintnet}, whose ground-equipment and facility vocabulary differs from OMIn aviation records. Overall token-F1 drops from 84.8\% to 46.4\% (Fig.~\ref{fig:transfer}). Component extraction remains 82.0\% ($-18$\,pp), while symptom extraction collapses to 12.6\% ($-85$\,pp). Yet end-to-end parser success remains at least 99.8\%. The contrast isolates two forms of portability: the output contract transfers because it is enforced at decoding time; field semantics do not transfer without target-domain vocabulary adaptation.

\subsection{Scope of the Claims}
Three results bound the contribution. First, the 97.3\% expert activation rate means the router provides explicit specialization control on OMIn, not sparse-expert efficiency. Second, NASD's full-data gain is 0.1\,pp; only the low-resource action-field effect is supported statistically. Third, 46.4\% MaintNet F1 is not production-ready cross-domain accuracy. ARID establishes an offline execution and validation path; a new maintenance domain still requires semantic adaptation.

\section{Conclusion}
ARID turns four-field maintenance extraction into a measured offline pipeline on an 8\,GB Jetson: 82.9\% deployed token-F1, 5{,}310/5{,}656\,ms resident P50/P99, 12.5\,W active power, and grammar-valid JSON. Its most useful finding is also its boundary: constrained decoding preserves the machine interface under domain shift, but it does not preserve semantic accuracy. Reliable deployment therefore requires both a stable edge execution path and target-domain adaptation; ARID provides the former and makes the latter visible.

\section*{Acknowledgment}
OpenAI IMAGE-2 was used to generate the visual draft of Fig.~1 from author-written prompts; the authors selected, corrected, and verified the final diagram. OpenAI Codex was used for language editing and consistency checks throughout the manuscript. The authors reviewed all AI-assisted output and remain responsible for the content.

\IEEEtriggeratref{7}
\bibliographystyle{IEEEtran}
\bibliography{references}

\end{document}